\documentclass{ecai}
\usepackage{times}
\usepackage{graphicx}
\usepackage{latexsym}
\usepackage{amsmath}
\usepackage{amsfonts}
\usepackage{amssymb}
\usepackage{subfig}

\usepackage{algorithm}
\usepackage[noend]{algpseudocode}
\usepackage{mathabx}
\usepackage{lmodern}
\usepackage[english]{babel}
\usepackage[autostyle, english = american]{csquotes}
\MakeOuterQuote{"}

\begin{document}

\title{Uniform State Abstraction For Reinforcement Learning}

\author{John Burden \institute{University of York, United Kingdom, jjb531@york.ac.uk} \and Daniel Kudenko \institute{L3S Research Center, Leibniz University Hannover, Germany }}

\maketitle
\bibliographystyle{ecai}

\begin{abstract}
Potential Based Reward Shaping combined with a potential function based on appropriately defined abstract knowledge has been shown to significantly improve learning speed in Reinforcement Learning. MultiGrid Reinforcement Learning (MRL) has further shown that such abstract knowledge in the form of a potential function can be learned almost solely from agent interaction with the environment. However, we show that MRL faces the problem of not extending well to work with Deep Learning. In this paper we extend and improve MRL to take advantage of modern Deep Learning algorithms such as Deep Q-Networks (DQN). We show that DQN augmented with our approach perform significantly better on continuous control tasks than its Vanilla counterpart and DQN augmented with MRL.
\end{abstract}

\section{Introduction}
Reinforcement Learning (RL) has gone through something of a renaissance in recent years, enjoying mainstream recognition with the success of DQN \cite{DQN} and Alpha-GO \cite{alphaGo}. RL agents principally learn through interaction with their environment. The goal for an agent is to learn a policy --- a map from states to actions --- that maximizes the expected cumulative reward during a single episode of interaction with a given environment. Due to the "Curse of Dimensionality" \cite{CoD}, it is infeasible to apply many "na{\"i}ve" RL approaches in anything but a very small number of dimensions --- the number of states explodes exponentially with the number of dimensions. This is an issue as basic RL algorithms such as Q-Learning typically need to visit a large number of the available states many times before a satisfactory policy is reached. 

Reward Shaping (RS) has proven to increase learning speed for RL agents \cite{shaping} at the cost of introducing some domain knowledge provided by an expert. The domain knowledge is encoded as an extrinsic reward function which is given to the agent in addition to its regular reward. 

However, encoding such domain knowledge is often expensive or infeasible. Initial attempts at automatically constructing the extrinsic reward function have been made \cite{graphRS}, but require the environment dynamics to be known a priori. 
An alternative approach relies on solving an abstract version of the task \cite{DomainExperts}\cite{plan}\cite{Marthi}. This still requires domain knowledge --- in the form of the abstract formulation of the task; however this is less knowledge than fully encoding the shaping function. Unfortunately, the task of encoding or identifying such knowledge 
for many applications is difficult and time consuming. This motivates research into automating further the process of constructing such knowledge. This could allow for reaping the benefits of using this knowledge without the onus of having to encode or even have the knowledge. 

One method for constructing abstractions uses conceptually simple uniform state aggregation combined with Potential-Based reward shaping  --- MultiGrid Reinforcement Learning (MRL) \cite{multigrid} . However, when this approach is extended to modern, deep learning approaches, the methods employed for learning a suitable potential function are insufficient. These insufficiencies arise from a high sample complexity for learning the values of abstract states when compared to the ground level. This results in a reward shaping function that often hinders learning. 

Our primary contribution is a method based on MRL which performs well in the deep learning setting. Our proposed method creates these abstractions by explicitly constructing Abstract Markov Decision Processes to create an abstraction of the task. This abstract task is easily solvable with dynamic programming methods and the solved values of abstract states provide a potential function for use with Potential Based reward shaping. The AMDP is constructed by uniformly partitioning each state-dimension into a pre-defined number of abstract cells. Exploration of the ground environment viewed through an abstract lens can allow us to build up an estimate of available abstract transitions. 

We compare our approach to both Vanilla DQN as well as DQN augmented with MRL. The three methods are evaluated in popular benchmark environments, Mountain Car, Puddle World and Catcher. We find empirically that our method outperforms the others significantly.

\section{Background}

\subsection{Reinforcement Learning}

Reinforcement Learning consists of interaction between an agent and an environment \cite{RLIntro}. The environment is usually represented by a Markov Decision Processes (MDPs) We use the standard notation for MDPs:
\[ 
\mathcal{M} = (S_\mathcal{M},A_\mathcal{M},R_\mathcal{M},P_\mathcal{M})
\]

Here $S_\mathcal{M}$ is a set of states in which the agent may find itself. $A_\mathcal{M}$ is a mapping from states $s \in S_\mathcal{M}$ to a set of actions available in $s$. $R_\mathcal{M}(s,a,s')$ is the reward function, detailing the immediate reward received by the agent when it transitions from state $s$ to $s'$ using action $a$. $P_\mathcal{M}(s,a,s')$ denotes the probability of reaching state $s'$ if currently in state $s$ and performing action $a$. The subscript $\mathcal{M}$ is used to allow us to distinguish between multiple MDPs and also between Abstract Markov Decision Processes which we introduce later. For a more complete overview of RL, see \cite{RLIntro}.

\subsection{Deep Q-Networks}
A widely used, modern  RL algorithm is Deep Q-Networks (DQN) \cite{DQN}. DQN is the "deep" expansion of Q-Learning \cite{RLIntro} and uses essentially the same update rules and operating principles as Q-Learning but adapted to use a neural network as its function representation.

As the agent interacts with its environment, the agent accumulates experience in the form of $(s, a, r, s')$ tuples, which are used to update the neural network. For each experience tuple, the values $Q(s,a)$ and $r + \max_{a'}{Q(s',a')}$ are calculated using the neural network. The difference between these two values is used in the loss function to update the network. This "bootstrapping" method enables the agent to learn strong estimates for the expected return of each state-action pair. 

Various alterations can be made to the "core" algorithm to improve performance. One of these is experience replay which disentangles chains of adjacent states and thus satisfies the iid assumption. Another common alteration is to use a target network, that is, to utilise a second network to compute the "target" $r + \max_{a'}{Q(s',a')}$. This second network has its weights copied from the first "predictor" network at regular intervals . Using a target network keeps the target stationary and makes it easier for the first network to converge.

DQN often suffers from a high sample complexity due to the aforementioned Curse of Dimensionality. Our method aims to reduce this by allowing the learning agent to evaluate its actions within the context of the "big picture", requiring fewer learning updates.

\subsection{Reward Shaping}
One method of giving knowledge to the agent is Reward Shaping (RS). Reward Shaping consists of giving an additional reward $F(s,a,s')$ to the agent after each transition. Intuitively, this additional reward represents external knowledge and is used to steer the agent towards more desirable behaviour.

It was shown in \cite{potentialRS} that if $F$ is defined as the difference of potential between two states, then the optimal policy learned by the agent will not change as a result of shaping. That is, we define 
\[ 
F(s,a,s') = \gamma\phi(s') - \phi(s)
\]

The primary issue to consider is the origin of the shaping function. Manually designing a function that indicates the quality of a transition or state is in many cases time consuming or infeasible. This motivates research into automatically constructing such a function.

\subsection{Reward Shaping with Abstract Markov Decision Processes} \label{PBRSAMDP}
One method for inducing a shaping function with relatively little external input is to solve an abstract version of the task. One way of 
defining abstract tasks is to use Abstract Markov Decision Processes (AMDPs). 

An AMDP represents a more abstract version of a ground-level MDP:
\[
\mathcal{A} = (S_\mathcal{A}, A_\mathcal{A}, R_\mathcal{A}, P_\mathcal{A})
\]

Each of the elements of a tuple correspond to abstractions of their counterparts in an MDP --- that is, they operate on abstract states in lieu of ground states. Defining these sets and functions is not trivial --- what abstract states should there be, what abstract actions and rewards etc ---- and they are often constructed utilising knowledge from a domain expert. 

In addition, an abstraction function $Z$ is required, mapping ground states $s \in S_\mathcal{M}$ to an abstract state $t \in S_\mathcal{A}$

When we have multiple agents operating on different levels of abstraction, we refer to the agent interacting with the MDP as the \textit{ground} agent and the agent interacting with the AMDP as the \textit{abstract} agent. Similarly, when focusing on learning on one of the decision processes we will denote this as abstract or ground learning.

There exist a number of examples using AMDPs together with Potential Based reward shaping \cite{DomainExperts}\cite{plan}\cite{Marthi}. Once the AMDP is constructed, dynamic programming is commonly used to compute a value $V$ for each state. The state value is then used for the potential function for shaping. That is, for state $s \in S_\mathcal{M}$, the potential function for shaping is set $\phi(s) \gets \omega V(Z(s))$, for some constant scaling variable $\omega$ and state abstraction function $Z$.

In both \cite{DomainExperts} and \cite{plan} the AMDPs were hand-crafted by an expert. In \cite{Marthi}, the AMDP is generated from an abstraction function $Z$ and set of Options --- macro actions --- $O$. If no set of Options is available, primitive actions can be used, but this limits the abstract actions to primitive actions, which introduces a lot of stochasticity into the AMDP. For the experimental domain given in \cite{Marthi} (Othello), constructing $Z$ involved knowledge about desirable positions of the game such as corners and edges. It also involved constructing a basic option to respond optimally within their abstract state-space.

There is an inherent trade-off to consider when using AMDPs for PBRS. The more "abstract" the AMDP is, the less useful the extrinsic reward will be for shaping, but the easier the AMDP will be to solve. The reverse also holds, not-so-abstract AMDPs will give very useful extrinsic reward but will be too costly to yield an appropriate value function for abstract states. The degree of abstraction will depend ultimately on the size of the task and the level to which abstract knowledge is available or can be obtained.

Whilst each of the mentioned approaches for shaping with AMDPs provide very good improvements to an agent's learning rate, they all require non-trivial external knowledge of the domain. It can be costly to encode such knowledge for many domains. Human error may also creep in and cause the agent to be directed away from rewarding behaviour. If incorrect knowledge is supplied, it can be remedied using knowledge revision \cite{KR}. However, this method can only correct the given abstract transition probabilities, it cannot amend information relating to the existence of abstract actions or states. 

\textit{Grze{\'{s}} and Kudenko}\cite{multigrid} give an approach known as MultiGrid Reinforcement Learning (MRL). This approach uniformly aggregates the MDP's state-space at two different resolutions. The coarser resolution is used to allow tabular RL on a continuous state-space. The finer-grained resolution is used to provide a set of abstract states. MRL then extends Sarsa by building up a value function for the abstract states concurrently with the $Q$ function using similar temporal difference updates. An externally provided reward is used for these updates. The value function over abstract states is then used as a potential function for PBRS as the agent enters new abstract states --- the same approach as the previously mentioned \cite{DomainExperts}\cite{plan}. Whilst this approach has been shown to be effective for the Mountain Car environment using state aggregation and classical (i.e. non-deep) Q-learning, it is completely out-classed by modern Deep RL algorithms. Further this method cannot simply be extended to utilise a neural network for the ground level representation --- for an agent utilising DQN, there is not a sufficient number of episodes to learn a useful abstract value function. Our approach extends this method in order to allow use with modern Deep RL algorithms --- this is achieved by explicitly constructing the AMDP and solving the abstract model that is created. We compare the results of our extension directly in our results for each domain in Section \ref{Results}. Section \ref{differences} describes the differences between our approach and MRL in greater detail.

\section{Method}
Imparting knowledge to an agent has shown itself to be a hugely useful paradigm for increasing the rate at which agents learn. Unfortunately it has also proven to be often intractable to obtain or encode this knowledge. Our method hopes to address this by reducing further the external knowledge required. 

In very broad strokes, our method is as follows:
First the set of abstract states is created using a uniform partitioning of the state-space across each state-dimension. 
A state abstract function is then created to map ground states to their corresponding abstract state. 

Secondly, the ground agent follows a pure exploration policy without making any updates to the neural network --- that is, we view the interactions through an abstract lens and only build up the AMDP's state transition function.

Once the allotted time for exploration has finished, the AMDP is constructed. This process uses the set of abstract states, and for each abstract state adds deterministic abstract actions representing each abstract transition that was observed. The abstract actions are deterministic, because for each observed transition a separate action is defined. The abstract reward is set to $-1$ for each transition in order to introduce a step penalty. The rationale for this is explained in Section~\ref{rewardConstruction} below. 

Finally, this AMDP is solved using Value Iteration and Potential Based reward shaping as described in Section \ref{PBRSAMDP}, using the value of the abstract states as the potential function.

We will now view each of these steps in more detail. We also give an overview of the AMDP construction and Value Iteration process in Algorithm \ref{alg}.

\begin{algorithm}
    \caption{AMDP Generation }
    \label{alg}
    \begin{algorithmic} 
        \Procedure{AMDP Generation} {MDP $M$, Partition List $D$, Exploration Policy $\pi_E$} 
        
            \State $S_\mathcal{M}$ = set of MDP states
            \State Partition $S_\mathcal{M}$ uniformly into $D_i$ bins for dimension $i$
            \State Denote these bins as elements of $\mathcal{S_\mathcal{A}} = (\bigtimes \mathcal{Z}_{D_i})$
            \State Create function $Z$ mapping state $s$ to bin containing $s$
            \State Initialize abstract transitions $P_\mathcal{A}(s,a,s') := 0$
            \State Initialize abstract Rewards $R_\mathcal{A}(s,a,s') := 0$
            \\
            
            \For{each exploration episode}
                \State $s :=$ Initial State
                \While{episode not complete}
                    \State Select action $a$ according to $\pi_E$
                    \State Perform action $a$ and observe $s', r$
                    \If $Z(s) \neq Z(s')$
                        \State $P(Z(s), Z(s')-Z(s), Z(s')) = 1$
                        \State $R(Z(s), Z(s')-Z(s), Z(s')) = -1$
                    \EndIf
                \EndWhile
            \EndFor
            
            \State return $(S_\mathcal{A}, A_\mathcal{A}, R_\mathcal{A}, P_\mathcal{A})$
          
        \EndProcedure
    \end{algorithmic}
\end{algorithm}

\subsection{Partitioning}
The first step to our method is to partition the state-space uniformly in each dimension. The only hyper-parameter to consider is the number of partitions per dimension. More partitions per dimension will allow a greater level of guidance for the agent over that dimension. The downside to using more partitions is an increased computational cost of solving the AMDP later on. For continuous state dimensions that have no upper or lower bound on the values we must also pick suitable upper and lower limits. Any encountered values outside of this range are set to partitions capturing any such values.

\subsection{Exploration}
Since there is no prior knowledge of the dynamics of the MDP, the AMDP cannot be built until we have explored the environment sufficiently. To do this, we allow the agent to follow an exploration policy, observing and recording the abstract transitions that are taking place. 

It is desirable to reduce the time spent on the exploration phase, in order to allow shaping to begin sooner. To achieve this, we do not perform any ground learning during the exploration phase. We simply focus on building up the AMDP by observing the ground agent's transitions through an abstract lens.

During the exploration phase, the agent will move between ground states and consequently abstract states. For exploration we also utilise a different level of states which we refer to as exploration states. We will revisit what exactly these exploration states are after introducing the exploration policy.

A dynamic exploration policy is given for the agent to follow. During this exploration, a visit count for each state-action pair is recorded. The agent picks actions in an epsilon greedy fashion with respect to this visit count, selecting with a higher probability the action with the lower visit count for the current state. This ensures that the agent tries to experience different states and actions, leading to a higher exploration rate.

For this exploration method, it turned out to be advantageous to use so-called exploration states instead of the AMDP states. Exploration states are simply a coarser abstraction of the state space than the AMDP's abstract states. This encourages the agent to explore areas of the environment that are further away than when using the finer-granularity AMDP abstract states. 
The resolution of the exploration state space was chosen to be very coarse and is given in the Hyper-parameters table \ref{table:hyperparams}.

\subsection{AMDP Construction}
Once the exploration phase is complete we can construct an AMDP representing an abstraction of the explored environment. 
The set of states $S_\mathcal{A}$ is the set of partitions we devised earlier. For each of these abstract states $t$, we create an abstract action set $A_\mathcal{A}(t) = \{ t':  t \rightarrow t' \text{ was observed} \}$. That is, if abstract transition $t \rightarrow t'$ was observed, then we give abstract state $t$ an action $t'$ representing the action that causes such a transition.

The abstract reward function deserves more discussion and is fully detailed in Section \ref{rewardConstruction}.

We also construct $P_\mathcal{A}$. This is rather simple. 
\[
P_\mathcal{A}(t,t') = \begin{cases}
1, \text{If transition } t \rightarrow t' \text{ was observed}\\
0, \text{otherwise}\\
\end{cases}
\]

Since we are only using the AMDP for shaping, and thus only utilising the shaped reward after an abstract transition has occurred, we can treat the abstract transitions as deterministic --- if the agent is in abstract state $t$, and performs abstract action $t \rightarrow t'$, it always transitions to $t'$. This is acceptable only because the agent is never actively making abstract decisions, only being rewarded for performing these actions when viewed through an abstract lens. 

\subsection{Constructing the AMDP Reward Function} \label{rewardConstruction}
Whilst the transition function detailing the dynamics of the environment viewed through our abstract states can easily be approximated through observation, defining a useful reward function has a few more possibilities. 

For an abstract transition $t \rightarrow t'$ one could opt to estimate the average observed cumulative reward over that transition period. However this may not be ideal for environments where there are many ways of achieving the abstract transition, with large variances in the cumulative ground reward observed. As a consequence, the abstraction is not able to distinguish between poor sequences of ground actions and a potentially optimal sequence, if the poor sequences are more frequent. This was also observed empirically, where using observed cumulative reward values worked well for environments such as Mountain Car, but performed poorly in environments where the ground reward can vary more, e.g. Puddle World.

A simpler alternative that we found to work well empirically was to set the abstract reward to $-1$ for every transition. Combined with the set of terminal abstract goal states, the value of an abstract state becomes the transition distance between the current abstract state and a goal. 
Despite our abstract states being the same size, this does not ensure that each abstract transition takes the same number of ground transitions --- particularly when some state-spaces incorporate agent velocity. 
This means that in some cases, where the agent moves through abstract states very slowly, that the AMDP will be rewarding non-optimal behaviour. Empirically, this doesn't seem to cause much of an issue --- the AMDP's extra reward and guidance being helpful \text{most} of the time is enough to improve performance. This supposed issue is exacerbated in the Catcher environment, where one of the state dimensions is velocity --- determining exactly how quickly the agent can move through the position-dimension abstract state. Despite this, we still see notable improvements in Catcher

\subsection{Abstract Goals}
Our method requires that one or more abstract states are selected as goal states. When the value function for the AMDP is being computed, the values of abstract goal states are set to $0$. This provides an endpoint for the step-penalty reward function. This usually does not require much additional knowledge of the domain ---  goals are often part of the task description or easy to describe.

Some environments require a single goal to be repeatedly achieved, for example the Catcher domain in Section \ref{catcher}. Our method handles this with no required alterations, repeatedly guiding the agent to the goal.

Our method can also be applied to environments which do not have conventional "goals" or domains with an infinite horizon. In such cases, the abstract goal can simply be set to "desirable" behaviour. This will then reward repeated completion of the abstract goal more-so than other behaviour.

\subsection{Exploitation}
Now that the appropriate AMDP is constructed, dynamic programming methods such as Value Iteration can be used to compute a value $V(s)$ for each abstract state $s$. We now can use our value function to augment an existing RL algorithm (We utilised DQN \cite{DQN}) using potential based reward shaping. Whenever the agent changes abstract state from $t \in S_\mathcal{A}$ to $t' \in S_\mathcal{A}$ we consider this an abstract transition $t \xrightarrow{} t'$. If such an abstract transition happens from our ground-level observations --- that is, we observe $s \xrightarrow{} s'$, such that $s \in t$, $s' \in t'$ --- the agent is given addition reward $F(t,t') = \gamma V(t') - V(t)$
Intuitively, this shaping rewards the agent for moving towards more promising abstract states (i.e., abstract states that have a higher potential).

The exploitation policy utilises an $\epsilon-greedy$ approach where the $\epsilon$ value is annealed over the course of the total number of episodes (for details see the hyper-parameter table \ref{table:hyperparams}. It is worth mentioning that this is an entirely different $\epsilon$ than is used in the exploration phase.

It is possible that during the exploration phase some abstract state $t$ is missed. If $t$ is then encountered in this exploitation phase, then the AMDP has no appropriate value $V(t)$. In this case, we can return a value $0$ and then resolve the AMDP with value iteration. The time to resolve the AMDP is very short due to the small size of the AMDP. In all of our test domains this did never occurred.

\subsection{Differences from MultiGrid Reinforcement Learning} \label{differences}
Since our method is based on MRL \cite{multigrid}, it is important to highlight where the two approaches differ. As previously stated, MRL can't simply be extended by changing the ground representation from state aggregation to a neural network --- the results in \ref{Results} show that this can lead to poor performance, the shaping function actually hindering learning. 

One large difference between our method and MRL is that our method explicitly builds up the AMDP, allowing us to use model-based methods (i.e., Value Iteration) in order to compute our shaping function. MRL on the other hand uses a temporal difference method to estimate the abstract values of its coarser state aggregates. MRL's approach works well when state aggregation is used for representing the ground level too as they have similar sample complexities. The ability of Deep Learning methods to generalise, however, causes MRL's state aggregation abstraction to "lag behind" the ground representation --- each coarse state aggregate needs visiting multiple times to yield an accurate abstract value. Our approach takes advantage of the explicit AMDP model by only requiring a state to be visited once in order to be factored in to the shaping function properly.

Building up the AMDP through exploration takes additional time in the form of an exploration phase. MRL does not require this due to its model-free approach. However, as shown in the results section (Section \ref{Results}), the exploration phase takes very little time. Further, we do not need to visit every single abstract state and observe every possible abstract transition --- just those that will appear in our final policy. 

MRL begins shaping immediately. Whilst this may appear beneficial (our method can only begin shaping after the exploration phase is finished and the AMDP has its abstract value function computed), PBRS with an ill-suited shaping function can hinder learning speed.


\section{Experimental Domains}
The three environments that were used were MountainCar, Continuous PuddleWorld and Catcher. These domains are all commonly used benchmark environments. Additionally, each domain has intuitive state transition dynamics, enabling easy conception of what a "good" abstraction may look like --- it is important that our chosen environments do not have optimal abstractions for AMDP-based PBRS that just happen to line up with our proposed uniform partitions. 

\begin{figure}
\center{
\subfloat[Mountain Car]{\includegraphics[width=0.147\textwidth]{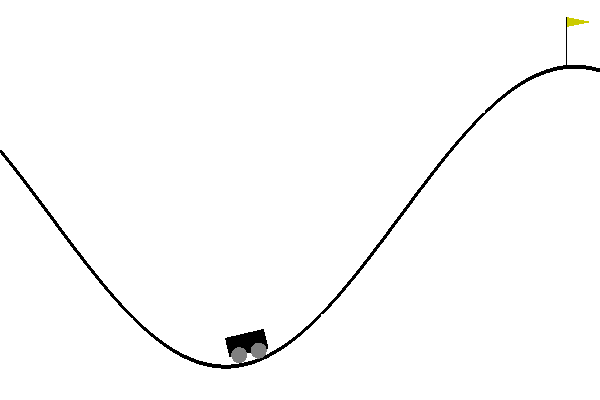}}
\hspace{0.01\textwidth}
\subfloat[Puddle World]{\includegraphics[width=0.147\textwidth, height=0.105\textwidth]{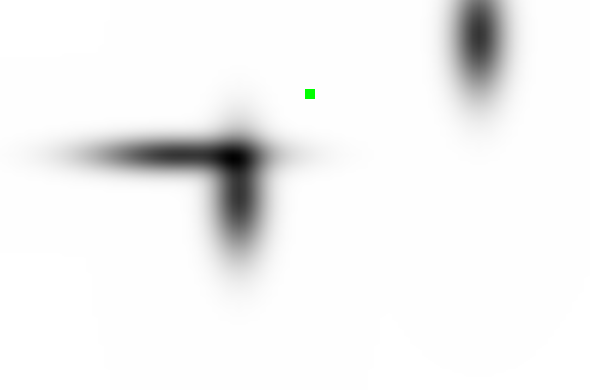}}
}
\hspace{0.01\textwidth}
\subfloat[Catcher]{\includegraphics[width = 0.105\textwidth]{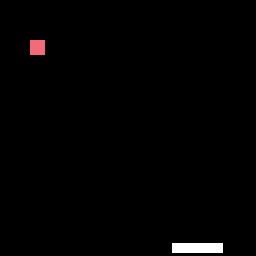}}
\caption{The environments used to evaluate our method}
\label{fig:Domains}
\end{figure}

Figure \ref{fig:Domains} shows a visual depiction of each environment. Briefly we will overview each one.

\subsection{Mountain Car}
In the Mountain Car environment, the agent is positioned somewhere inside a valley and must reach the top of the right hand side. The state-space consists of the x-position and velocity of the car. The agent has three actions, $left$, $neutral$ and $right$, indicating an amount of force to apply in the designated direction. A reward of $-1$ is received after each action the agent makes. An episode terminates upon reaching an $x$-position of $0.5$ or after $200$ steps have elapsed. It  is important to note that the car cannot reach the top of the hill simply by moving to the right --- the car needs to build up momentum first by swinging back and forth.

\subsection{Continuous Puddle World}
In the Continuous Puddle World environment the agent is situated on a two-dimensional plane, ranging on values from $(0,0)$ to $(1,1)$. The agent begins in the bottom left quadrant and must reach very close to $(1,1)$. There are $5$ discrete actions available to the agent; the agent may move in any cardinal directions (by an randomly determined, but bounded amount), as well as standing still. There is a puddle occupying certain spots in the plane (shown in the depiction). The agent receives a reward of $-1$ for each step, as well as additional negative reward based on how deep into the puddle the agent is. Ideally we want the agent to move to the top right corner receiving as large a reward as possible. An episode terminates on reaching a satisfactory distance to $(1,1)$ or if $250$ steps elapse. 

\subsection{Catcher} \label{catcher}
In the catcher game, the agent embodies a one dimensional, horizontal line. Small squares fall from above the agent, perpendicular to the agent's axis of movement. This agent has three actions, $left$, $neutral$ and $right$ which moves the agent in the corresponding direction. The agent's goal is to move itself to intercept the falling square. 
The state-space consists of the agent's $x$ position and velocity, as well as the square's $x$ and $y$ position. For each square the agent intercepts, it receives a reward of $1$. For each square that it misses it receives a reward of $-1$. After $3$ misses in total, the episode ends. The episode also terminates after $500$ steps in order to prevent episodes becoming inordinately long as the agent improves. This allows the collection of approximately $15$ balls. 

\section{Experimental Details and Results} \label{Results}
We now show our empirical results and give an analysis. 
Our proposed method was evaluated in the three environments, Mountain Car, Puddle World and Catcher. We augment a DQN method using our method and compare it against both an unaugmented DQN agent and DQN augmented with MRL as a baseline. The hyper-parameters for each agent and environments are in table \ref{table:hyperparams}. 
The shared hyper-parameters were chosen empirically in order to optimise the performance of the baseline DQN agent. The hyper-parameters that were only used by our method were chosen empirically in order to maximise the performance against time. 

For MRL in each environment, the sum of the ground rewards received during the transition between coarse aggregations was used to update the coarse value function, as defined in the original paper.

The abstract reward function used for Mountain Car was the sum of received ground rewards during the abstract transition. This was chosen to enable an easier comparison with MRL (Mountain Car was the only domain used for evaluation in the original MRL paper \cite{multigrid}). For both Puddle World and Catcher we used an abstract reward function of $-1$ for each abstract transition as this was found to perform better for our approach.

\begin{table*}
\begin{center} 
    \begin{tabular}{|c|c|c|c|} 
        \hline
        Parameter & Mountain Car & Puddle World & Catcher \\
        \hline
        $\alpha$ & $1e-3$ & $5e-4$ & $1e-5$\\
        $\gamma$ & $0.995$ & $0.99$ & $0.95$\\
        $\tau$ & $1e-2$ & $1e-2$ & $1e-2$\\
        $\omega$ & $1$ & $1$ & $1$\\
        $\epsilon$ & $0.1 \rightarrow 0.01$ & $0.2 \rightarrow 0.05$ & $0.1 \rightarrow 0.01$ \\
        Abs. Size & $(50,50)$ & $(50,50)$ & $(20,10,20,10)$\\
        Exp. Size & $(5,5)$ & $(5,5)$ & $(10,5,10,5)$\\
        Episodes & $500$ & $1000$ & $1000$\\
        Exp. Episodes & $500$ & $1000$ & $500$\\
        Action Rep. & $64$ & $64$ & $16$ \\
        
        \hline
    \end{tabular}
\end{center}
\caption{Hyper-parameters used for experiments}
\label{table:hyperparams}
\end{table*}

We show our empirical results for each domain. In each of the below results, the mean reward of each agent is plotted, with confidence intervals of $95\%$ shaded.  Our augmented agent is shown in grey against the unaugmented DQN agent in blue and DQN augmented with MRL in green.
Since our approach initially follows an exploration policy to construct the AMDP - it uses more episodes than the others. For a fair comparison, we opt to compare the agents by plotting reward against elapsed time --- including the exploration phase. This also fairly accounts for any extra computation our method uses to solve the AMDP or compute the extrinsic reward. The end of the exploration phase is indicated by the vertical dotted black line. We ran the three agents for a set number of episodes, meaning the agents varied in the amount of time taken to complete the task. Since we are more interested in comparing performance against time, the shortest time taken to complete all of the episodes was taken as a benchmark and the other agents had their results truncated to that time-step.
Comparing the agents episodically directly is somewhat misleading, this is because our method completes a lot of episodes very quickly during the exploration phase. However if we offset for the exploration episodes we see similar results to those shown for time --- we omit these for space.

\begin{figure}
\center{
\subfloat[Mountain Car]{\includegraphics[width = 0.45\textwidth]{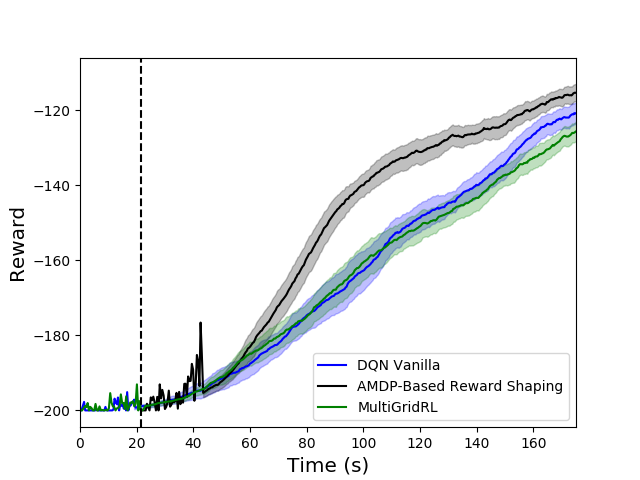}\label{MCResults}}\\
\subfloat[Puddle World]{\includegraphics[width=0.45\textwidth]{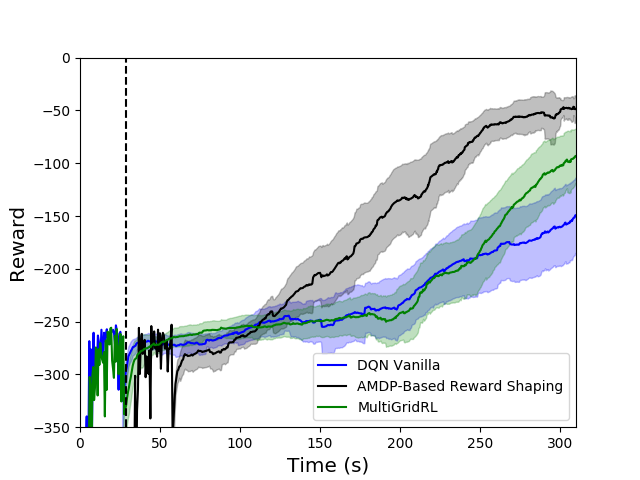}\label{PWResults}}
}\\
\center\subfloat[Catcher]{    \includegraphics[width = 0.45\textwidth]{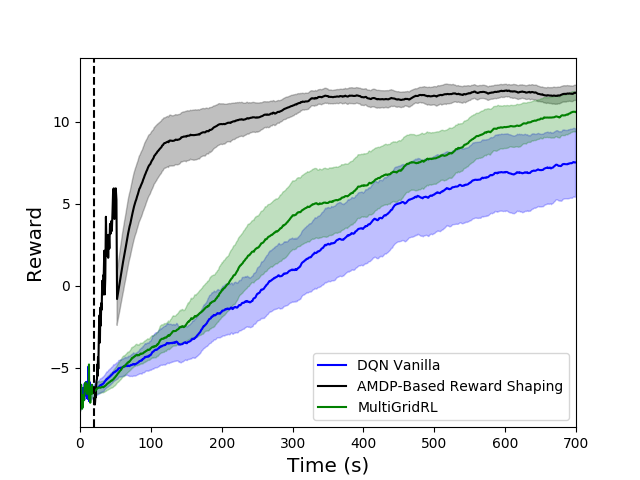}\label{CResults}}

\caption{Results comparing each agent's performance across each domain. The mean value of episodic reward is plotted against the time at which the episode terminated. The curves have been smoothed using a moving average of size 50 for visual clarity. The first 50 episodes are plotted without smoothing or averaging to show the results before the moving average can be calculated --- this explains the apparent drop in performance early on in the learning process. The shaded regions indicate a 95\% confidence interval.}
\label{fig:Results}
\end{figure}

In each domain we can see that our approach beats both other approaches with a significant margin. The exploration phase takes very little time and gives a large performance boost very quickly. MRL is more of a mixed-bag, improving the learning performance of DQN mildly in Catcher and PuddleWorld, while somewhat hindering it Mountain Car. It is notable however that in none of the environments did MRL out perform our approach.

Overall these results show that MRL is ill-suited for Deep Learning due to the large amount of interactions required for learning its abstract value function. The results further show that our extension of MRL is an improvement for the Deep Learning setting, and that DQN augmented with our approach significantly outperforms Vanilla DQN, without the need to provide much domain knowledge.

\section{Related work}
Here we overview related work in the area of generating and utilising knowledge to improve agent performance. There are two main areas that we consider, Hierarchical-based approaches and Option discovery. 
\subsection{Hierarchical Structures}
Other RL methods exploit hierarchical structures in the same spirit as we do to achieve impressive results. Hierarchical Deep Q-Networks (HDQN) \cite{hdrl} uses two networks, one producing abstract goals and rewards for the second to try to achieve. HDQN achieves impressive results on the infamous Montezuma's Revenge, although requiring a pre-trained object detection algorithm to identify possible sub-goals. 

Deep Abstract Q-Networks (DAQN) \cite{daqn} utilise AMDPs in order to produce sub-goals for their ground agent to achieve. Their AMDP takes propositional properties about the state of the game as their state-space and constructs sub-tasks for the ground agent to achieve. This method was evaluated on a toy version of Montezuma's revenge and performed very well in that domain, as well as performing well in the "single-life" version of the real Montezuma's revenge game. It is notable however, that this method had hand-crafted abstract state-space as well as giving the agent extra information such as which room it is currently in. This extra information and state-space represents a large amount of domain knowledge given to the agent. 

A major difference between both HDQN and DAQN and our proposed method is the mechanism for utilising the abstraction. HDQN and DAQN both yield agent control to the abstract agent by allowing the agent to proactively select sub-goals and ground policies --- our method, on the other hand, influences the agent by providing extra reward for desirable abstract behaviour. The AMDP, whilst instrumental in forming the extrinsic reward function, is never given control over the ground agent's actions. The agent's behaviour is influenced by the abstraction re-actively. The proactive nature of HDQN and DAQN allows them to take advantage of abstract knowledge immediately, whereas the reactive nature of our method lets us treat all AMDPs as deterministic --- the observed transitions have already occurred. This makes it easier for our method to construct useful AMDPs. Overall both approaches, proactive and reactive, have their pros and cons.

\subsection{Option Discovery}
Options are an extension to MDPs to allow Semi-Markov behaviour. An Option is essentially a chain of primitive actions from the MDP based on a low-level policy, along with predefined invocation and termination policies \cite{options}. It has been shown that having useful Options available can improve learning performance \cite{options}. However, Each Option typically needs to be provided by a domain expert and requires knowledge of how to achieve the sub-task each Option embodies. As with constructing useful AMDPs, defining Options manually is often prohibitively time consuming. 

Efforts have been made to generate Options automatically. The most common method for Option discovery has been to identify bottlenecks in the state-space and learn Options to manoeuvre between such bottlenecks--- the logic being that bottlenecks provide clear sub-tasks for the agent to complete. Research has focused on different methods for identifying bottlenecks, such as constructing clusters for state-reachability \cite{bottleneck}, employing many ant-like agents to drop pheromones and search for areas of high traffic \cite{ants}, utilising Spectral Clustering \cite{evenMoreBottlenecks}\cite{moreBottlenecks} to create clusters and bottlenecks, as well as creating clusters based on entropy measures of the MDPs known transition graph \cite{bottleneckEntropy}.

These approaches vary in their efficacy and generality, many of the approaches struggle to identify bottlenecks of a non-uniform width throughout the environment or even just bottlenecks of size greater than one. 

Other than bottleneck methods, other Option generation methods exist, such as approaches that utilise Extended Sequence Trees (ESTs) \cite{EST}\cite{EST2}. These EST-based methods create seemingly more nuanced Options, but potentially struggle with ESTs which grow exponentially with the number of actions available in each state.  

Another method uses Association Rule Mining \cite{arm} to generate Options but require the sub-goals identified a-priori --- this method constructs options than complete these sub-goals in an optimal ordering. This still requires a lot of domain knowledge about what constitutes a desirable sub-goal and how to formulate them.

\section{Conclusion And Outlook}
Our principal contribution has been to show that uniform partitions of state-spaces can be utilised to generate simple AMDPs which are easily solved. 

We did this by utilising a separate exploration phase that built up the AMDP based upon these uniform partitions. Once these AMDPs were sufficiently constructed, value iteration is used to produce a value function for each abstract state. PBRS then utilised these values for its potential function for use with shaping. 

For all of our three empirical evaluation domains our method improved the performance when compared against DQN. This approach demonstrates the efficacy of both generating and utilising knowledge based on a learning agent's experiences.

A core aspect of our method is solving the AMDP, and if the AMDP is too large due to a very high dimensional state-space we may not be able to find a solution in a sufficiently short time frame. In very high dimensional tasks, convolutional layers are used to reduce the number of state dimensions, but this can not be directly applied to AMDP states. Possible solutions include utilising more expert knowledge to identify "key" dimensions to focus on, or introducing a non-uniformity to the state partitions that possibly cross dimensional lines. Another potential extension would be to utilise convolution layers and perform the abstraction after the salient features have been identified. 

\bibliography{ecai}

\begin{thebibliography}{10}

\bibitem{CoD}
R.~Bellman, Rand Corporation, and Karreman Mathematics~Research Collection,
  {\em Dynamic Programming}, Rand Corporation research study, Princeton
  University Press, 1957.

\bibitem{DomainExperts}
Kyriakos Efthymiadis and Daniel Kudenko, `A comparison of plan-based and
  abstract mdp reward shaping', {\em Connection Science}, {\bf 26}(1),  85--99,
  (2014).

\bibitem{KR}
Kyriakos Efthymiadis and Daniel Kudenko, `Knowledge revision for reinforcement
  learning with abstract mdps', in {\em Proceedings of the 2015 International
  Conference on Autonomous Agents and Multiagent Systems}, AAMAS '15, pp.
  763--770, Richland, SC, (2015). International Foundation for Autonomous
  Agents and Multiagent Systems.

\bibitem{ants}
Mohsen Ghafoorian, Nasrin Taghizadeh, and Hamid Beigy, `Automatic abstraction
  in reinforcement learning using ant system algorithm',  9--14, (01 2013).

\bibitem{arm}
Behzad Ghazanfari and Matthew~E. Taylor, `Autonomous extracting a hierarchical
  structure of tasks in reinforcement learning and multi-task reinforcement
  learning', {\em CoRR}, {\bf abs/1709.04579}, (2017).

\bibitem{EST}
Sertan Girgin, Faruk Polat, and Reda Alhajj, `Improving reinforcement learning
  by using sequence trees', {\em Machine Learning}, {\bf 81}(3),  283--331,
  (Dec 2010).

\bibitem{plan}
M.~Grzes and D.~Kudenko, `Plan-based reward shaping for reinforcement
  learning', in {\em 2008 4th International IEEE Conference Intelligent
  Systems}, volume~2, pp. 10--22--10--29, (Sept 2008).

\bibitem{multigrid}
Marek Grze{\'{s}} and Daniel Kudenko, `Multigrid reinforcement learning with
  reward shaping', in {\em Artificial Neural Networks - ICANN 2008}, eds.,
  V{\'e}ra K{\r{u}}rkov{\'a}, Roman Neruda, and Jan Koutn{\'i}k, pp. 357--366,
  Berlin, Heidelberg, (2008). Springer Berlin Heidelberg.

\bibitem{bottleneck}
Ghorban Kheradmandian and Mohammad Rahmati, `Automatic abstraction in
  reinforcement learning using data mining techniques', {\em Robotics and
  Autonomous Systems}, {\bf 57}(11),  1119 -- 1128, (2009).

\bibitem{evenMoreBottlenecks}
Ramnandan Krishnamurthy, Aravind~S. Lakshminarayanan, Peeyush Kumar, and
  Balaraman Ravindran, `Hierarchical reinforcement learning using
  spatio-temporal abstractions and deep neural networks', {\em CoRR}, {\bf
  abs/1605.05359}, (2016).

\bibitem{hdrl}
Tejas~D. Kulkarni, Karthik Narasimhan, Ardavan Saeedi, and Joshua~B. Tenenbaum,
  `Hierarchical deep reinforcement learning: Integrating temporal abstraction
  and intrinsic motivation', {\em CoRR}, {\bf abs/1604.06057}, (2016).

\bibitem{bottleneckEntropy}
Shie Mannor, Ishai Menache, Amit Hoze, and Uri Klein, `Dynamic abstraction in
  reinforcement learning via clustering', in {\em Proceedings of the
  Twenty-first International Conference on Machine Learning}, ICML '04, pp.
  71--, New York, NY, USA, (2004). ACM.

\bibitem{graphRS}
M.~Marashi, A.~Khalilian, and M.~E. Shiri, `Automatic reward shaping in
  reinforcement learning using graph analysis', in {\em 2012 2nd International
  eConference on Computer and Knowledge Engineering (ICCKE)}, pp. 111--116.
  IEEE, (Oct 2012).

\bibitem{Marthi}
Bhaskara Marthi, `Automatic shaping and decomposition of reward functions', in
  {\em Proceedings of the 24th International Conference on Machine Learning},
  ICML '07, pp. 601--608, New York, NY, USA, (2007). ACM.

\bibitem{DQN}
Volodymyr Mnih, Koray Kavukcuoglu, David Silver, Alex Graves, Ioannis
  Antonoglou, Daan Wierstra, and Martin~A. Riedmiller, `Playing atari with deep
  reinforcement learning', {\em CoRR}, {\bf abs/1312.5602}, (2013).

\bibitem{potentialRS}
Andrew~Y. Ng, Daishi Harada, and Stuart~J. Russell, `Policy invariance under
  reward transformations: Theory and application to reward shaping', in {\em
  Proceedings of the Sixteenth International Conference on Machine Learning},
  ICML '99, pp. 278--287, San Francisco, CA, USA, (1999). Morgan Kaufmann
  Publishers Inc.

\bibitem{shaping}
Jette Randl{\o}v and Preben Alstr{\o}m, `Learning to drive a bicycle using
  reinforcement learning and shaping', in {\em Proceedings of the Fifteenth
  International Conference on Machine Learning}, ICML '98, pp. 463--471, San
  Francisco, CA, USA, (1998). Morgan Kaufmann Publishers Inc.

\bibitem{daqn}
Melrose Roderick, Christopher Grimm, and Stefanie Tellex, `Deep abstract
  q-networks', {\em CoRR}, {\bf abs/1710.00459}, (2017).

\bibitem{alphaGo}
David Silver, Aja Huang, Chris~J. Maddison, Arthur Guez, Laurent Sifre, George
  van~den Driessche, Julian Schrittwieser, Ioannis Antonoglou, Veda
  Panneershelvam, Marc Lanctot, Sander Dieleman, Dominik Grewe, John Nham, Nal
  Kalchbrenner, Ilya Sutskever, Timothy Lillicrap, Madeleine Leach, Koray
  Kavukcuoglu, Thore Graepel, and Demis Hassabis, `Mastering the game of {Go}
  with deep neural networks and tree search', {\em Nature}, {\bf 529}(7587),
  484--489, (jan 2016).

\bibitem{RLIntro}
Richard~S. Sutton and Andrew~G. Barto, {\em Introduction to Reinforcement
  Learning}, MIT Press, Cambridge, MA, USA, 2 edn., 2017.

\bibitem{options}
Richard~S. Sutton, Doina Precup, and Satinder Singh, `Between mdps and
  semi-mdps: A framework for temporal abstraction in reinforcement learning',
  {\em Artificial Intelligence}, {\bf 112}(1),  181 -- 211, (1999).

\bibitem{moreBottlenecks}
Nasrin Taghizadeh and Hamid Beigy, `A novel graphical approach to automatic
  abstraction in reinforcement learning', {\em Robotics and Autonomous
  Systems}, {\bf 61}(8),  821 -- 835, (2013).

\bibitem{EST2}
E.~Çilden and F.~Polat, `Toward generalization of automated temporal
  abstraction to partially observable reinforcement learning', {\em IEEE
  Transactions on Cybernetics}, {\bf 45}(8),  1414--1425, (Aug 2015).

\end{thebibliography}
\end{document}